\documentclass[10pt,twocolumn,letterpaper]{article}

\usepackage{cvpr}
\usepackage{times}
\usepackage{epsfig}
\usepackage{graphicx}
\usepackage{amsmath}
\usepackage{amssymb}
\usepackage[toc,page]{appendix}

\usepackage{float}
\usepackage{color}
\usepackage[linesnumbered, boxed, ruled]{algorithm2e}
\usepackage{multirow}
\usepackage[pagebackref=true,breaklinks=true,letterpaper=true,colorlinks,bookmarks=false]{hyperref}
\usepackage{array}
\newcolumntype{C}[1]{>{\centering\let\newline\\\arraybackslash\hspace{0pt}}m{#1}}
\usepackage{booktabs}
\usepackage{subfigure}

\usepackage{authblk}
\cvprfinalcopy 

\def\cvprPaperID{****} 

\begin{document}
\title{HAKE: Human Activity Knowledge Engine}
\author{Yong-Lu Li, \quad Liang Xu, \quad Xinpeng Liu, \quad Xijie Huang, \quad Yue Xu, \quad Mingyang Chen,\quad  Ze Ma,\quad  Shiyi Wang,\quad  Hao-Shu Fang,\quad  Cewu Lu*\\
	Shanghai Jiao Tong University\\
	\url{http://hake-mvig.cn}\\
	{\tt\small yonglu\_li@sjtu.edu.com liangxu@sjtu.edu.cn xinpengliu0907@gmail.com huangxijie1108@gmail.com\\}
	{\tt\small silicxuyue@gmail.com cmy\_123@sjtu.edu.cn maze1234556@sjtu.edu.cn Shiy.Wang@outlook.com fhaoshu@gmail.com lucewu@sjtu.edu.cn}
}
\maketitle
\renewcommand{\thefootnote}{\fnsymbol{footnote}}
\footnotetext{Draft, work in progress. *Cewu Lu is the corresponding author.}
\begin{abstract}
Human activity understanding is crucial for building automatic intelligent system. With the help of deep learning, activity understanding has made huge progress recently. But some challenges such as imbalanced data distribution, action ambiguity, complex visual patterns still remain. To address these and promote the activity understanding, we build a large-scale Human Activity Knowledge Engine (HAKE) based on the human body part states. Upon existing activity datasets, we annotate the part states of all the active persons in all images, thus establish the relationship between instance activity and body part states. Furthermore, we propose a HAKE based part state recognition model with a knowledge extractor named Activity2Vec and a corresponding part state based reasoning network. With HAKE, our method can alleviate the learning difficulty brought by the long-tail data distribution, and bring in interpretability. Now our HAKE has more than \textbf{7 M+} part state annotations and is still under construction. We first validate our approach on a part of HAKE in this preliminary paper, where we show \textbf{7.2} mAP performance improvement on Human-Object Interaction recognition, and \textbf{12.38} mAP improvement on the one-shot subsets.
\end{abstract}

\begin{figure}[!ht]
	\begin{center}
		\includegraphics[width=0.45\textwidth]{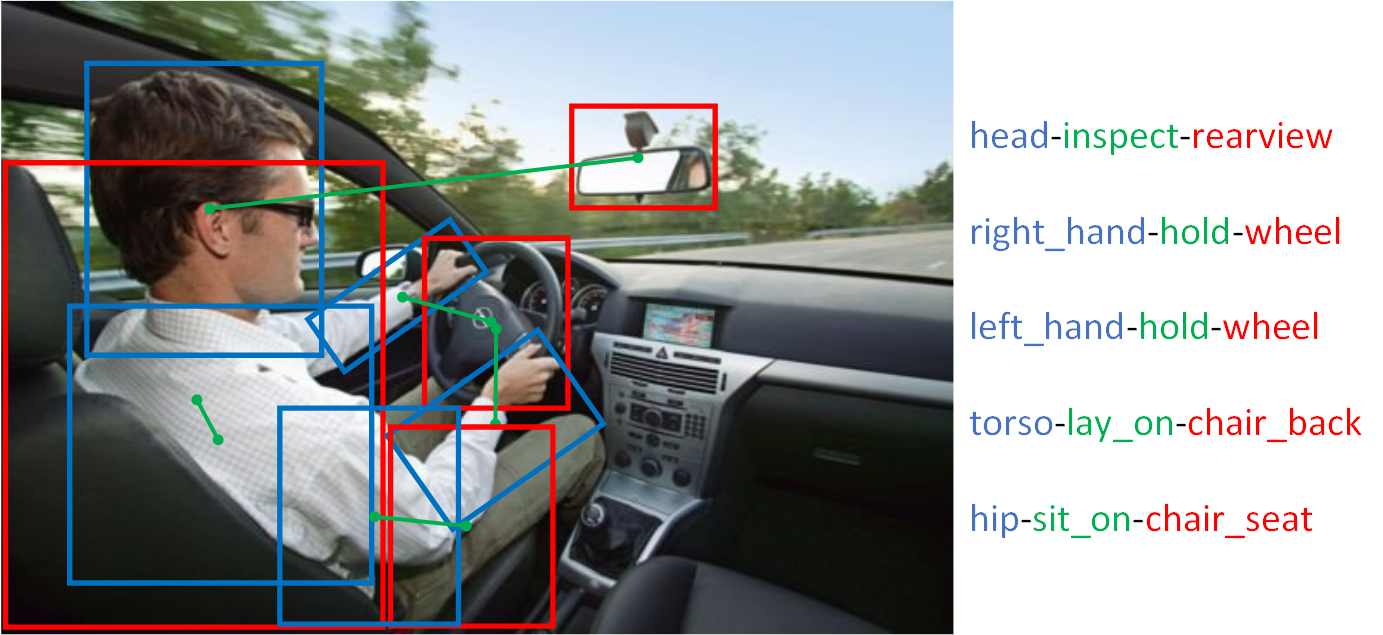}
	\end{center}
	\caption{Body part state samples. A driving scene contains activity $\langle person, drive, car \rangle$. It can be decomposed into various body part states like $\langle head, inspect, rearview\rangle$, $\langle right\_hand, hold, wheel\rangle$, $\langle left\_hand, hold, wheel\rangle$.}
	\label{Figure:ps_example}
\end{figure}

\begin{figure*}[!ht]
	\begin{center}
		\includegraphics[width=1\textwidth]{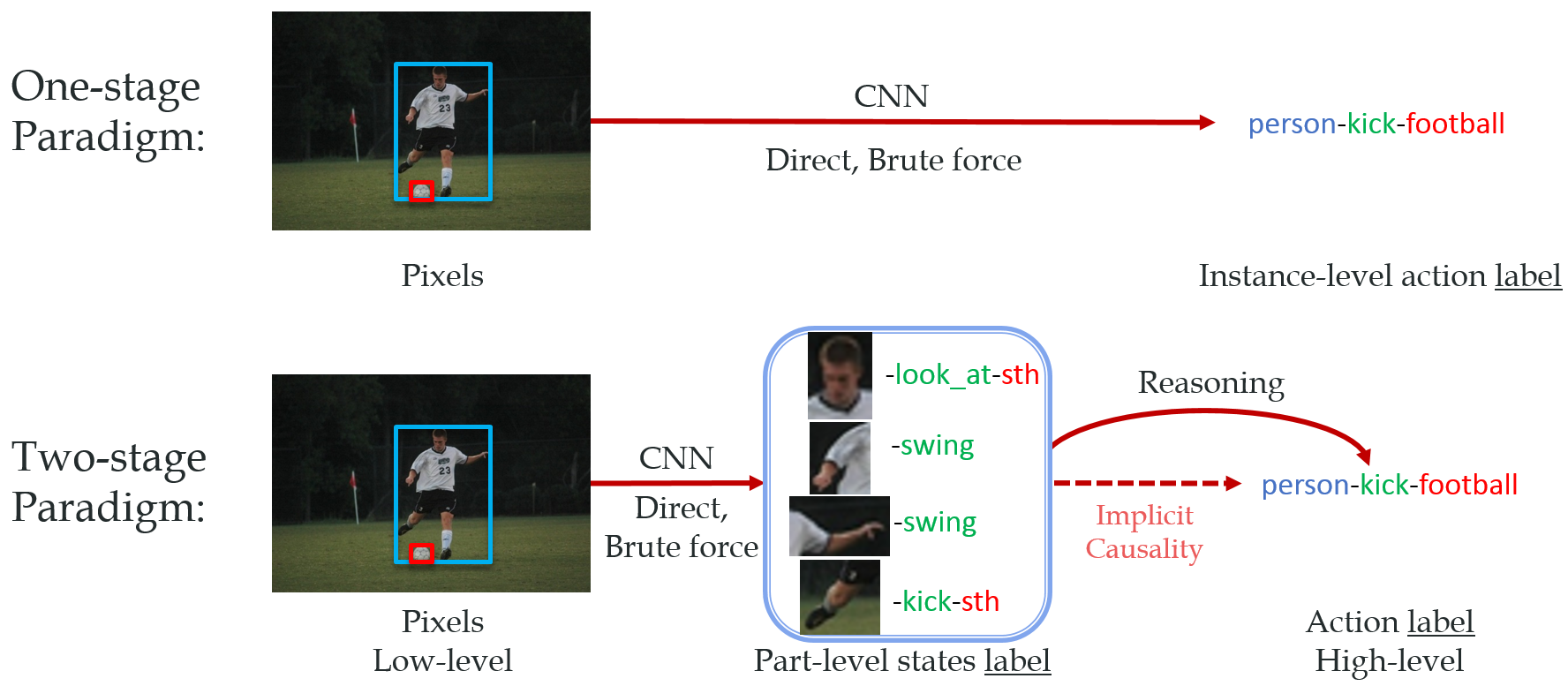}
	\end{center}
	\caption{Previous one-stage paradigm and our hierarchical two-stage paradigm.}
	\label{Figure:paradigm_compare}
\end{figure*}

\section{Introduction}
Human activity understanding is an active topic in computer vision and has a large number of potential applications and business prospects. Facilitated by the growth of image and video data and the renaissance of Deep Neural Networks (DNNs), lots of works have been proposed to forward this direction. Activity recognition has strong relations with other research contents of computer vision, such as object detection\cite{faster-rcnn}, pose estimation~\cite{fang2017rmpe}, video analysis~\cite{activitynet}, visual relationship~\cite{Lu2016Visual}.
Recent works on activity and action recognition almost rely on the end-to-end supervised paradigm to address this high-level cognition task, \ie perception from raw pixels directly to the activity classes in one stage. This paradigm shows poor performance on large-scale activity benchmarks, such as HICO~\cite{hico}, HICO-DET~\cite{hicodet}, AVA~\cite{AVA}. 

The limited performance of present one-stage paradigm on these large-scale and exceedingly difficult datasets are possibly due to the own difficulties of activity understanding. For instance, activity recognition has many challenges such as long-tail data distribution, variability and complexity of action visual patterns, crowd background in daily scenes, various camera viewpoints and motions, occlusion and self-occlusion, crowd-sourced annotations and data.
In the absence of data, one-stage paradigm which needs to bridge the huge gap is powerless. To this end, we propose a new Human Activity Knowledge Engine (HAKE) based on body part states~\cite{lu2018beyond}. Based on HAKE, a new corresponding hierarchical two-stage paradigm for activity recognition is also presented. 

\begin{figure*}[!ht]
	\begin{center}
		\includegraphics[width=1\textwidth]{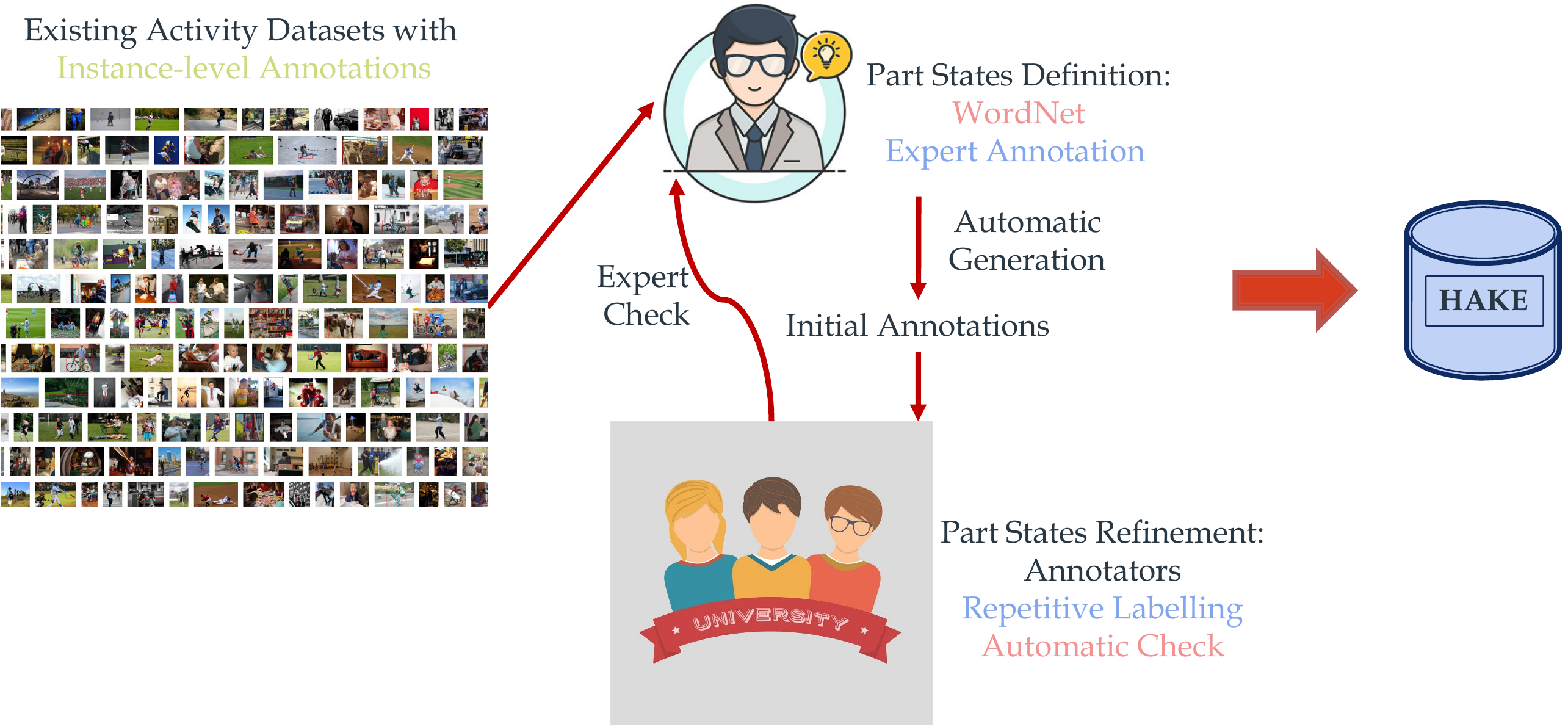}
	\end{center}
	\caption{The construction of our HAKE.}
	\label{Figure:model}
\end{figure*}

Different from the one-stage paradigm, we divide the activity understanding into two phases: 1. Part states~\cite{lu2018beyond} recognition from the visual patterns, and the activity representation by combining visual and linguistic knowledge; 2. Reasoning the activities from part states, as seen in Fig. \ref{Figure:paradigm_compare}.
Part states mean the finer level atomic body part actions~\cite{lu2018beyond} which compose the action of human instance. Fig. \ref{Figure:ps_example} shows an example of instance activity and its corresponding part states.
Based on the reductionism~\cite{honderich2005oxford}, our assumption is that: the human instance action consists of the atomic actions or states of all the body parts. Thus we can divide the instance action recognition into two sub-tasks: body part state recognition and the recombination of part states. 

The most obvious advantages of our hierarchical paradigm are three-fords.
First, part states are the basic components of instance actions, their relationship can be in analogy with the amino acid and protein, letter and word. Different instance actions like ``person hold an apple'' and ``person eats an apple'' share the same part states ``hand holds something'' and ``head looks at something''. 
Thus the imbalanced data problem will be greatly alleviated, for that the samples per category largely increases on the same data scale. For supervised learning, it will effectively reduce the learning difficulty. Furthermore, part state recognition is much easier than the instance action recognition because of fewer categories and simpler visual patterns. In our experiment, a simple model consists of shallow CNNs and fully connected layers can achieve acceptable performance on part state recognition, which is generally relative 50\% higher than the instance action recognition.
Second, with part state recognition as the midpoint, the gap between the image space and the semantic space would be greatly narrowed.
Third, we can obtain a more powerful representation of action patterns based on part states. In our experiment, the combinative visual-linguistic part state embeddings present obvious semantic meaning and better interpretation. When the model predicts what he/she is doing, we can easily know the reasons: what his/her body parts are doing.

The main contributions of this work are:
1. We construct a large-scale Human Activity Knowledge Engine named HAKE that bridges the relationship between instance activity and body part states. We will keep on enlarging and enriching it, and call on the community to help us make it more powerful to promote activity understanding.
2. A new hierarchical paradigm is proposed based on HAKE, which outperforms state-of-the-art methods on several activity recognition benchmarks. In particular, the performance of rare action categories on several benchmarks are significantly boosted.


\section{Construction of HAKE}
In this section, we will illustrate the construction of HAKE. Considering the complexity, we first construct part states annotations on still images, and then expand to the consecutive frames of videos. The key characteristic of our HAKE are: the definition of part states are based on atomic and composite actions, crowd-sourced images from several widely-used activity datasets, realistic visual contexts, diversity and variability of activities. 

\noindent{\bf Part States Definition.} HAKE is based on the existing well-designed datasets, for example, HICO-DET~\cite{hicodet}, V-COCO~\cite{vcoco}, OpenImage~\cite{openimages}, HCVRD~\cite{hcvrd}, HICO~\cite{hico}, MPII~\cite{MPII}, AVA~\cite{AVA}, which are structured around a rich semantic ontology. The activity categories contained in HAKE are chosen according to the most common human daily actions/activities, social interactions with daily objects and person. 
We first select 154 instance activity categories from above datasets in the case of hierarchical activity structure~\cite{activitynet}. All the part states will be annotated upon instance level activities.
Then we decompose the human body into ten body parts following~\cite{Fang2018Pairwise}, namely head, arms, hands, hip, legs, feet.
Third, we select about 200 part states based on the verbs from WordNet~\cite{miller1995wordnet} as the candidates to build a part state pool, e.g. ``hold'', ``push'', ``pick'' for hands, ``listen to'', ``eat'', ``talk to'' for head and so on. 

To ensure the quality of part state selection, we invite several experts to use their own understandings to depict the selected 154 instance actions in the body part level. For example, when we show an image with activity ``person drive a car'' to them, they may describe it as ``hip sit on something'', ``hands hold something'', ``head look at something''. Based on their choices, we use the Normalized Point-wise Mutual Information (NPMI)~\cite{church1990word} to calculate the co-occurrence between the instance action categories and part state candidates. Finally, we choose 92 candidates with the highest NPMI values as the final part states.

\begin{figure}[!ht]
	\begin{center}
		\includegraphics[width=0.45\textwidth]{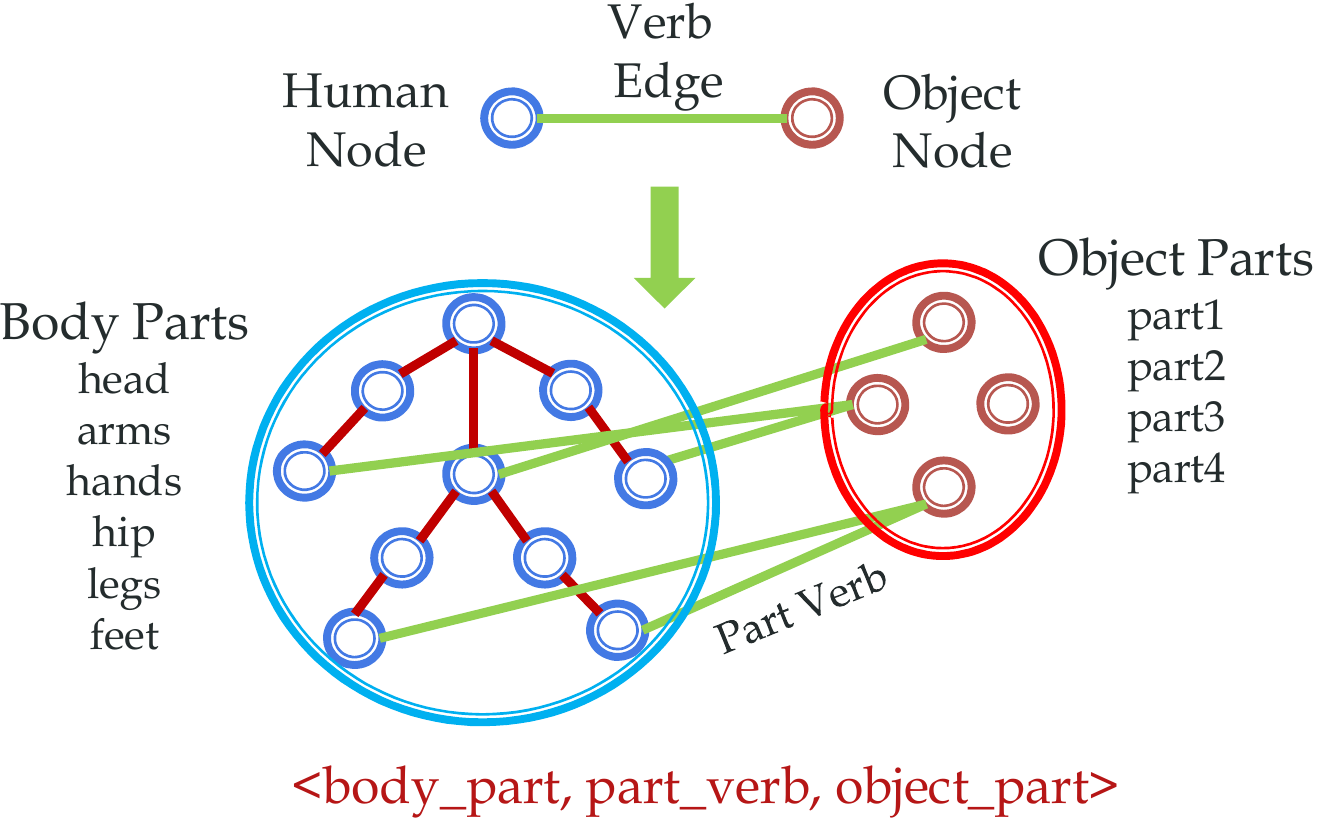}
	\end{center}
	\caption{The graphic model of instance activity and part states.}
	\label{Figure:graph}
\end{figure}

Based on 154 instance activities and 92 part states, we can construct a hierarchical graph structure shown in Fig.\ref{Figure:graph}. Actions and part states are represented as the edges between the subject and object nodes (for the non-object actions, the edge is a loop). 

\noindent{\bf Part State Annotation.} We annotate all part states belonging to all the actions of all the active persons in all the collected images exhaustively. To be specific, existing datasets already have the human and object bounding box annotations and the relationship links between them. We then use pose estimation~\cite{fang2017rmpe} to obtain the pose keypoints of all the annotated persons, and their part bounding boxes following~\cite{Fang2018Pairwise}. 
We adopt a semiautomatic method to build HAKE. First, we invite nine experts to annotate 10 thousands of images with all the 154 instance actions as the basis, and generate the initial part states for all the rest of images based on their annotation distribution. 
Thus the other annotators will use our tool to amend and refine the initial annotations according to their understanding of these actions. To ensure the quality, one instance with multiple actions would be annotated multiple times for each activity. Furthermore, each image will be checked at least twice by the automatic procedure and experts. We cluster these labels and discard the obvious outliers to obtain the robust label agreements. 

It is worth noting that, action recognition is a multi-label classification problem, an active person may have more than one actions. For each instance action, we annotate its corresponding part states respectively and then combine all sets of part states in the final round. In other words, a body part can also have multiple states at the same time, e.g., activity ``person cuts an apple'' would have part states ``right-hand holds a knife'' and ``right-hand uses something to cut something'' at the same time.

At present, we have finished the annotations of 104 K+ images, which include 677 k+ human instances, 278 K+ interacted objects, 733 K+ instance actions, 7 M+ human body part states. 
In addition, our labeling is still in progress, and we have build a project page (\url{http://hake-mvig.cn}) and an online annotation tool. We hope the volunteers from all over the world to help us enlarge and enrich HAKE to advance the activity understanding. With these densely annotated part states, we believe in that we can deepen and promote the activity understanding significantly.

\begin{figure}[!ht]
	\begin{center}
		\includegraphics[width=0.45\textwidth]{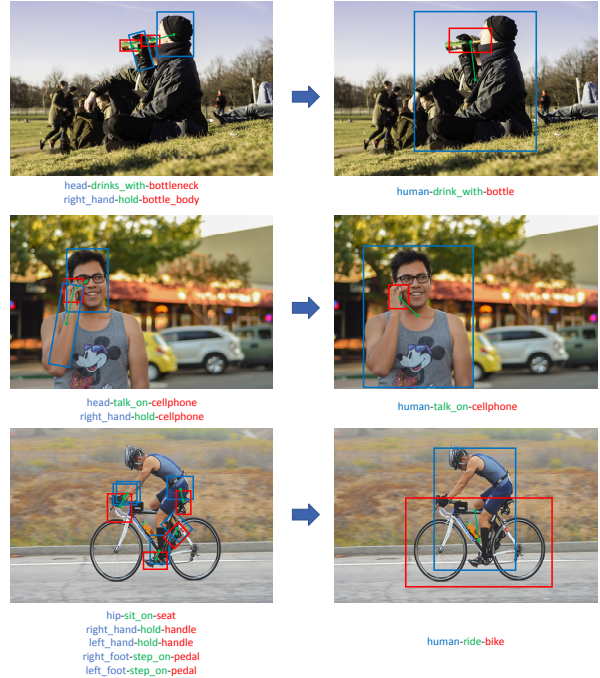}
	\end{center}
	\caption{Samples of activity reasoning via part states. By combining the classified part states, we can reason out the activity from co-occurrence and prior knowledge. Inner relations between part states can also be helpful. For instance, holding an apple by hand and eating an apple with mouth often appear together. These two part states can derive the activity $\langle human, eat, apple\rangle$.}
	\label{Figure:reasoning}
\end{figure}

\section{Hierarchical Paradigm}
On the basis of our human activity knowledge engine, we can address the activity recognition in a hierarchical way: 1. Part State Recognition with knowledge extraction via Activity2Vec; 2. Reasoning from Part States to Instance Activity. This hierarchy would bring in more interpretability and a new possibility for the following-up researches.

\subsection{Part State Recognition and Activity2Vec}
\begin{figure*}[!ht]
	\begin{center}
		\includegraphics[width=1\textwidth]{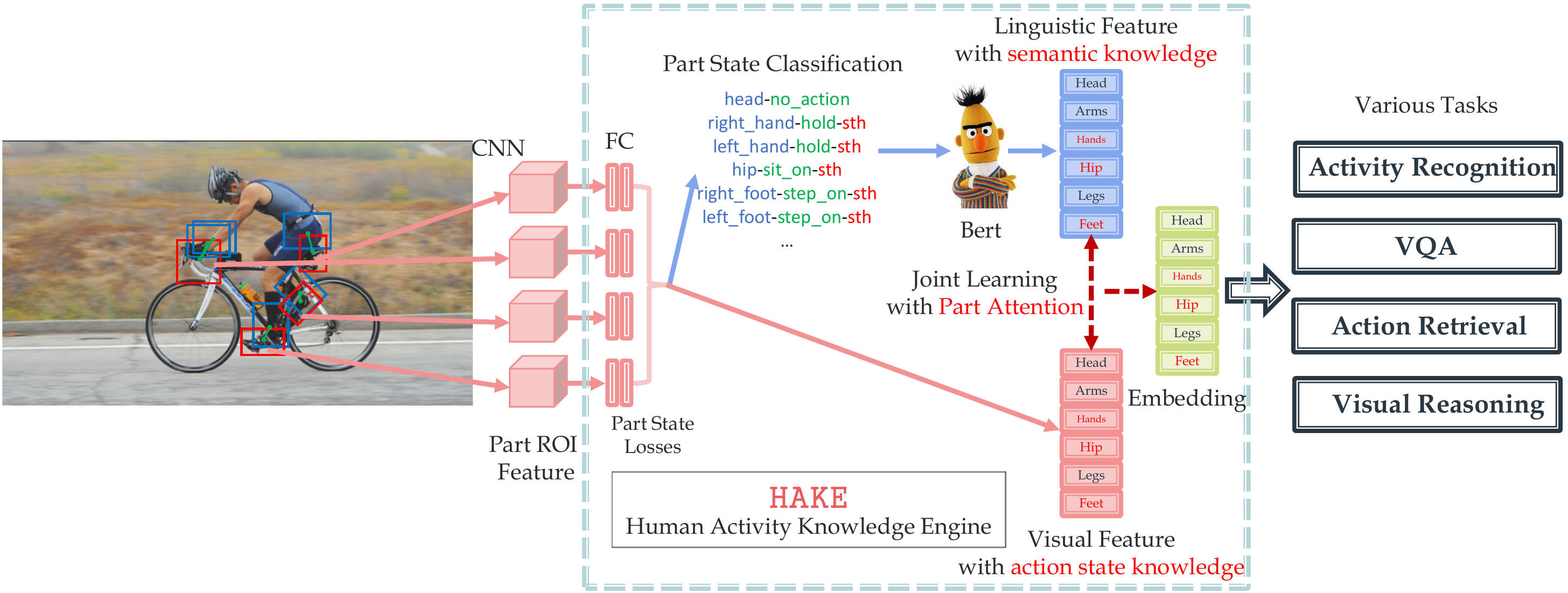}
	\end{center}
	\caption{The overview of Part state recognition and Activity2Vec.}
	\label{Figure:overview}
\end{figure*}

In the first phase, we utilize the canonical pipeline to address the part state recognition. With the object detection of images, we can obtain the bounding boxes of all the detected person and objects.
Second, we extract the ROI pooling features of all body parts as the input of the Part States Classification Network (PSC), as shown in Fig.~\ref{Figure:overview}. Within HAKE, we have annotated all the part states of all the human instances, thus we can construct part state classification loss for each part. As mentioned before, part state recognition is much easier in the supervised learning paradigm, which is also proven by our experiments.
Besides, we also use part-level interactiveness predictors~\cite{li2018transferable} for all parts, which can infer the relationship between each body part and the object. If a body part has interactions with the object, thus its interactiveness label will be one. These interactiveness labels are also converted from the part states labels. More details can be found in~~\cite{li2018transferable}. With the part-level interactiveness predictions, we can obtain the \textbf{body part attentions}, which indicate whether a part is important for the action recognition. For example, in the action ``person eat an apple'', only the hands and head are essential for the action classification. All the visual body part features will be multiplied with interactiveness attentions first.

To enhance the representation ability and promote the subsequent activity reasoning, we additionally utilize the uncased BERT-Base pre-trained model~\cite{devlin2018bert} as the language feature extractor. 
Bert~\cite{devlin2018bert} is a language understanding model trained on large-scale word corpus, it can generate contextual embeddings for many types of downstream NLP tasks. Bert has considered the surrounding (left and right) of a word, and use a deep bidirectional Transformer to extract the general embeddings of words. Thus its pre-trained representations carry the contextual information from the enormous text data, \eg Wikipedia. These features usually contain implicit semantic knowledge about the activity and part states, which is clearly different from visual information. 
In specific, we divide a part state into tokens, \eg ``head''(body part), ``eat''(verb), ``apple''(object). Each part state will be converted to a 2304 vector (three 768 vectors for the part, verb, object respectively). Second, we multiply the fixed language features with predicted part state probabilities from the part state classifiers, which can also be seen as an attention mechanism. A more possible part state will get a larger attention.

Our goal is to bridge the gap between part states and instance activity. The combination of the visual and linguistic knowledge thus can be a powerful clue for establishing this mapping. We align the visual and linguistic features by using triplet loss~\cite{schroff2015facenet}, and concatenate them as the output, this process is called as Activity2Vec. (Seen in Fig.~\ref{Figure:overview}) The output embedding is 3584 sized and as the input of the downstream tasks, \eg activity recognition, Visual Question Answering, action retrieval. Especially, before utilizing this embedding, we will first use the activity recognition task to pre-train it to capture the activity knowledge. 
From the experiment results, we can find that the embedding generated by our Activity2Vec can significantly improve the performance of multiple activity related benchmarks. 

\subsection{Reasoning from Part States to Instance Activity}
With the embeddings from the Activity2Vec, we can better infer the relationship between part states and activities. If all the activities can be seen as the nodes at a higher level within a hierarchical graph, the part states will be the nodes in the lower level. Inner relationships between part state nodes can be seen as their co-occurrence, so do the activity nodes. The edges linked the instance activity and part states are the key elements in activity understanding, as seen in Fig.~\ref{Figure:reasoning}. 

We propose a Part States Reasoning Network (PSR) to estimate these cross-level edges between activity nodes and part state nodes. In our vanilla version, we directly use the fully-connected layers to infer the actions from the combined part-level features, which obtains surprising improvements.
More details of the proposed PSC and PSR models will be illustrated in our official version paper.
\begin{figure*}[!ht]
	\begin{center}
		\includegraphics[width=1\textwidth]{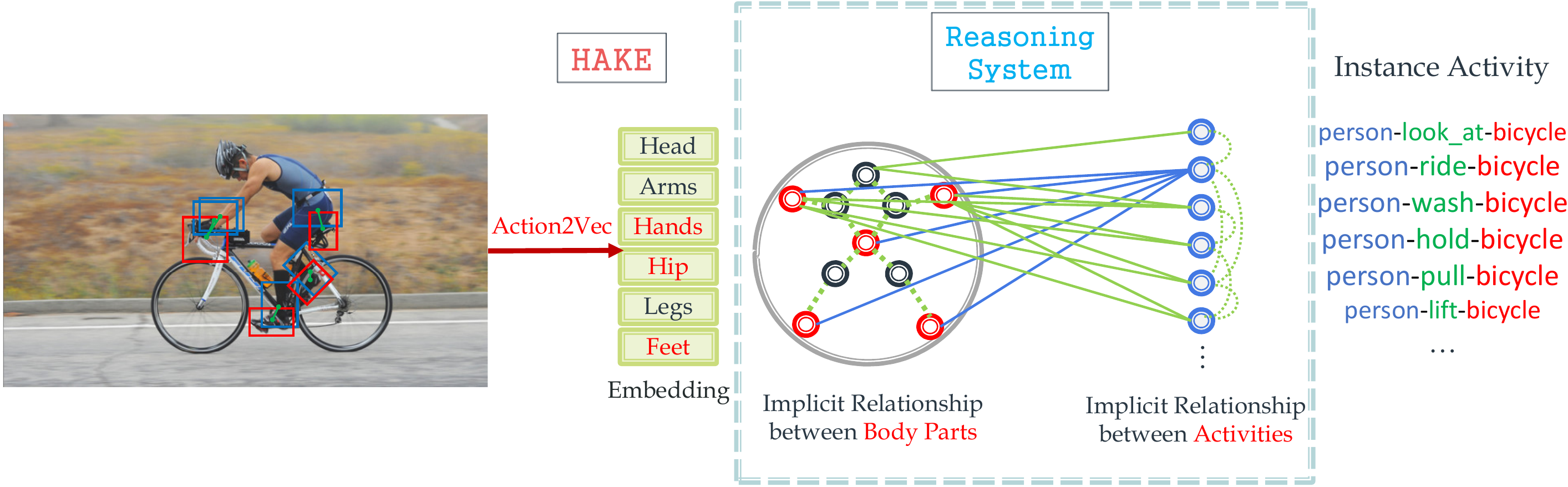}
	\end{center}
	\caption{Reasoning from part states to instance activities.}
	\label{Figure:reasoning}
\end{figure*}

\section{Experiments}
\subsection{An analogy: simplified Action Recognition}
\begin{figure}[!ht]
	\begin{center}
		\includegraphics[width=0.45\textwidth]{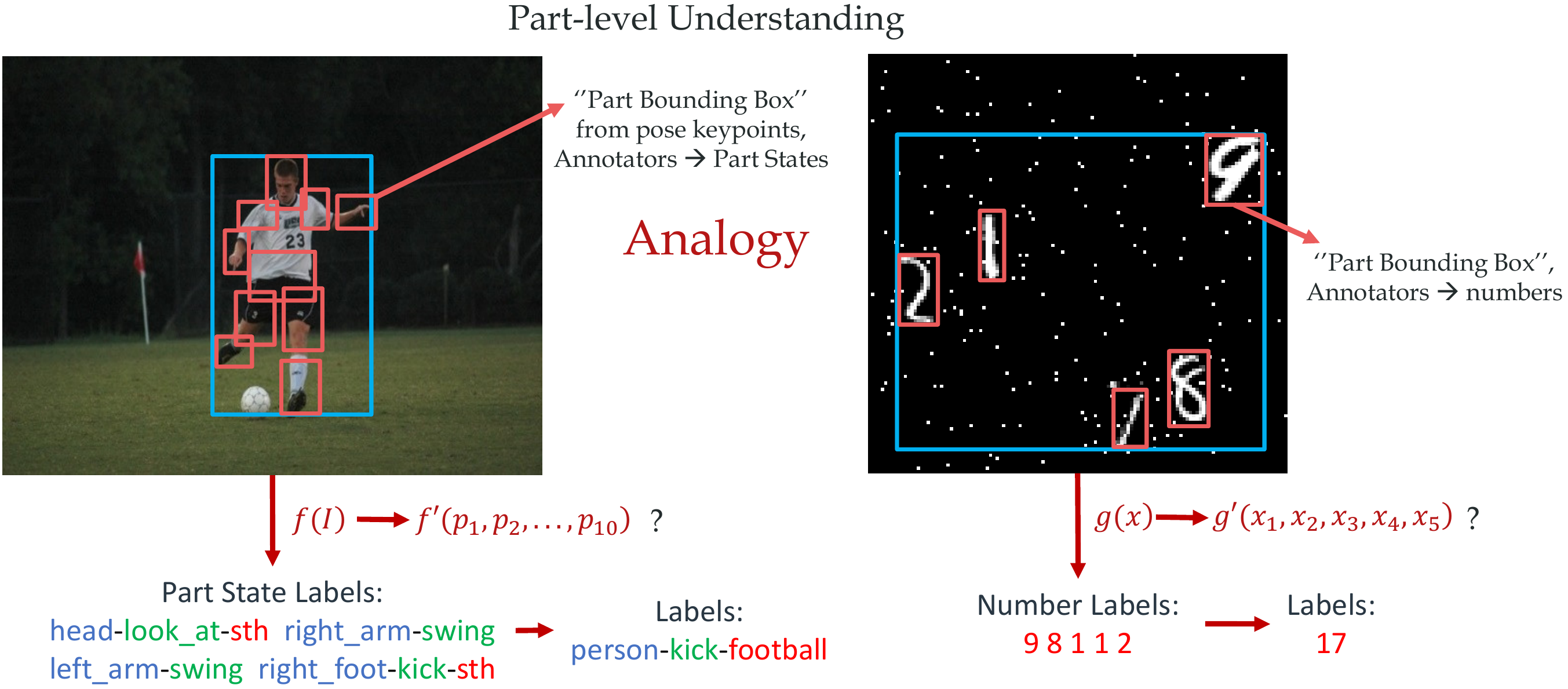}
	\end{center}
	\caption{An analogy to activity recognition.}
	\label{Figure:analogy}
\end{figure}
In this section, we design a simplified experiment to give a better intuition about our approach. We build a dataset derived from MNIST which consists of handwritten digits from 0 to 9 and with size 28~$\times$ 28~$\times$ 1. 
To test our assumption, we generate a set of new 128~$\times$ 128~$\times$ 1 images which is a combination of 3 to 5 handwritten digits randomly selected from MNIST and randomly distributed in images. And the corresponding label of each new image is the sum of the largest and second largest value of the digits within this image. Thus the total number of categories is 19 (0 to 18).
This problem is a simplified analogy of human activity recognition, as shown in Fig.~\ref{Figure:analogy}. We argue that actions can be decomposed into a set of part states, which highly resembles the relationship between the digits and the sum function in the above case. The random amount of digits and their distribution aim to simulate the randomness of part states. Taking the influence of background into consideration, we also add Gaussian noise on the whole generated image samples.

\begin{figure}[!ht]
	\begin{center}
		\includegraphics[width=0.45\textwidth]{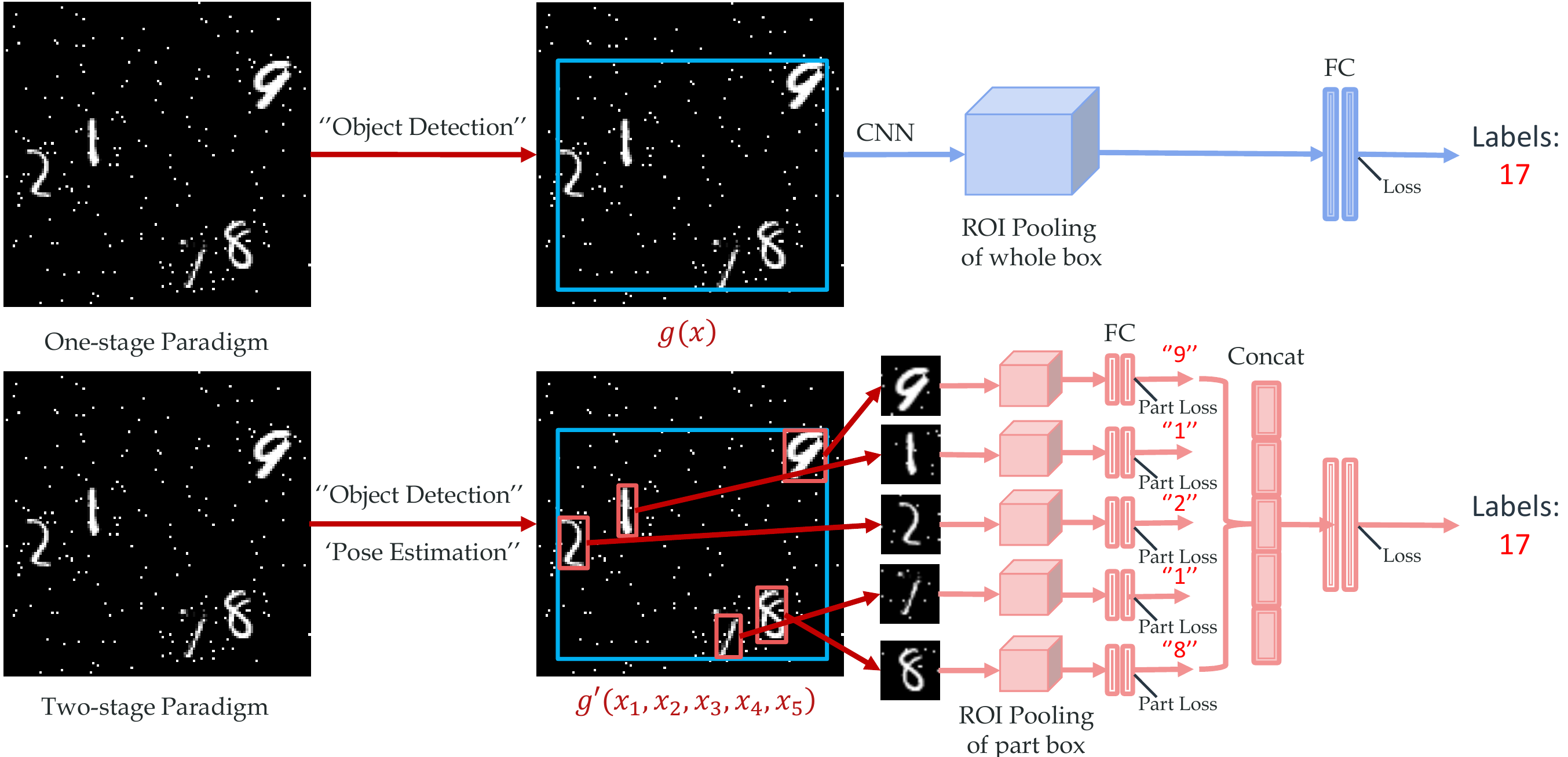}
	\end{center}
	\caption{Instance based model and part based model.}
	\label{Figure:mnist}
\end{figure}

To compare the one-stage (instance based) paradigm and two-stage (part based) paradigm, we adopt a simple network to conduct a test. The network is composed of shallow sequential convolution layers and fully connected layers (shown in Fig.\ref{Figure:mnist}). 

\begin{figure} [!ht]
  \centering 
  \subfigure[Loss]{ 
    \label{Figure:exp1_loss} 
    \includegraphics[width=0.23\textwidth]{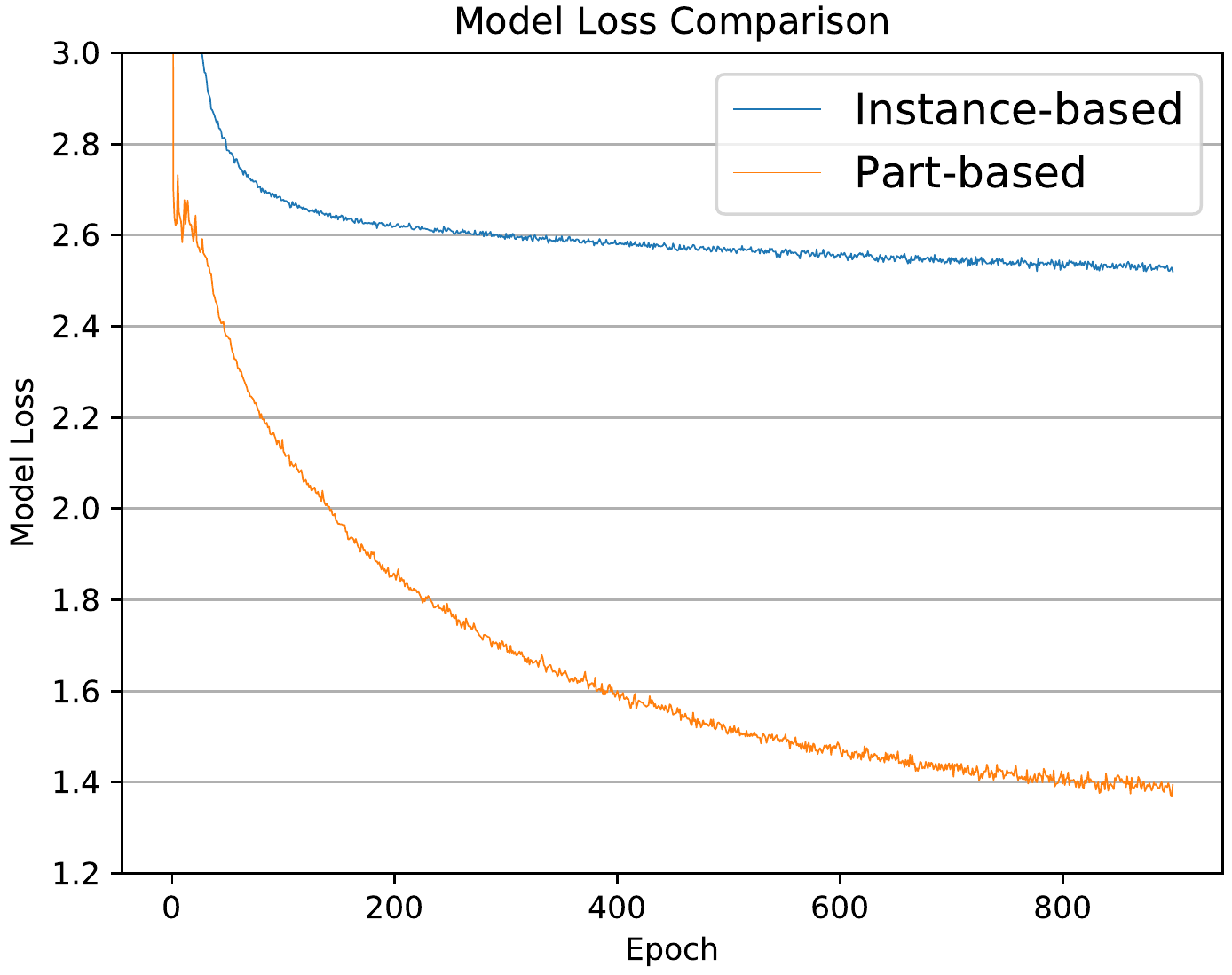}} 
  \subfigure[Test Accuracy]{ 
    \label{Figure:exp1_acc} 
    \includegraphics[width=0.23\textwidth]{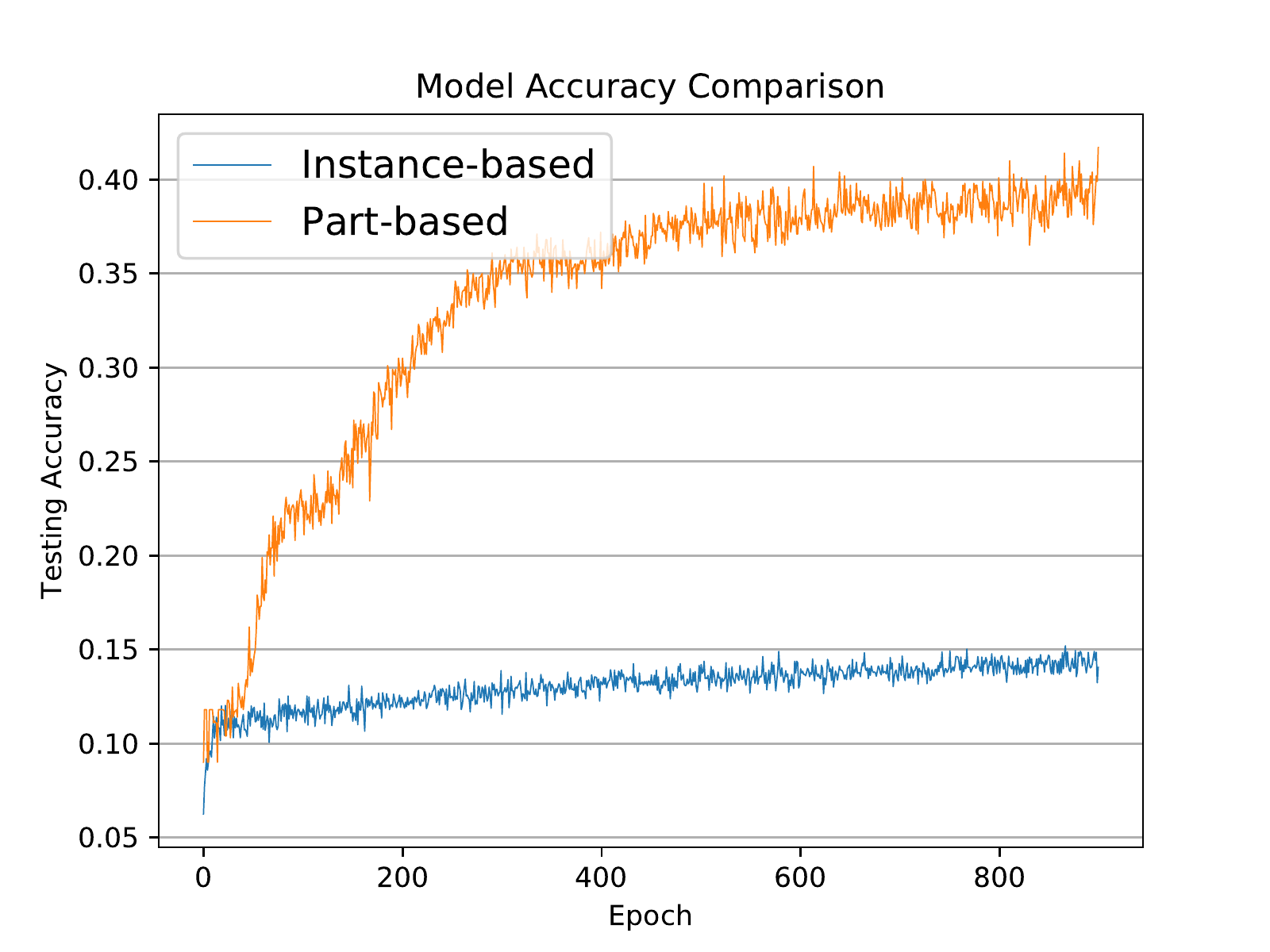}} 
  \caption{Comparison of loss and accuracy} 
  \label{Figure:loss_acc} 
\end{figure}

Two paradigms are trained with the same optimizer, learning rate and epochs. The results are shown in Fig.\ref{Figure:loss_acc} and Tab.~\ref{table:analexp}, which show the significant superiority of part based paradigm (174\% relative increases on accuracy) over instance based paradigm. To some extent, this result supports our assumption about the decomposability of human instance activity and the effectiveness of part state knowledge representation.

\subsection{Human-Object Interaction Recognition}
To verify the effectiveness of our HAKE and hierarchical paradigm, we perform experiments on Human-Object Interaction (HOI) recognition task~\cite{hico} in still images. HOI~\cite{hicodet,gao2018ican,li2018transferable} usually accounts for a large proportion of daily life activities. And it is more complex than the non-HOI activity, \eg ``dance'', ``swim''.
In the initial version, we just report the results on HICO~\cite{hico} to evaluate the improvements brought by HAKE, especially on one-shot and few-shot learning problems. 
More results on other activity understanding tasks will be reported in the official paper. 

\begin{table}[!ht]
	\begin{center}
		\begin{tabular}{cc}
		\hline  
	    Method & Test Accuracy \\
	    \hline  
	    Instance Based Paradigm& 15.2  \\
	    Part Based Paradigm & \textbf{41.7}  \\
	    \hline
		\end{tabular}
	\end{center}
	\caption{Comparison of accuracy on our dataset}
	\label{table:analexp}
\end{table}
HICO~\cite{hico} contains 38,116 images in train set and 9,658 images in test set. 
\if Above the annotations on the image level, we enhance the labels with the body part states. For example, an image is not merely labeled with multiple HOIs, but also several body part states. Fig.\ref{Figure:part-state} has shown some annotations. \fi 
To be fair, we follow the experiment settings of~\cite{Fang2018Pairwise} to compare the recognition performance. Considering that our HAKE-based part-level representation is complementary to the instance-level representation, we use the late fusion strategy to generate the final prediction. 
\begin{figure*}[!ht]
	\begin{center}
		\includegraphics[width=1\textwidth]{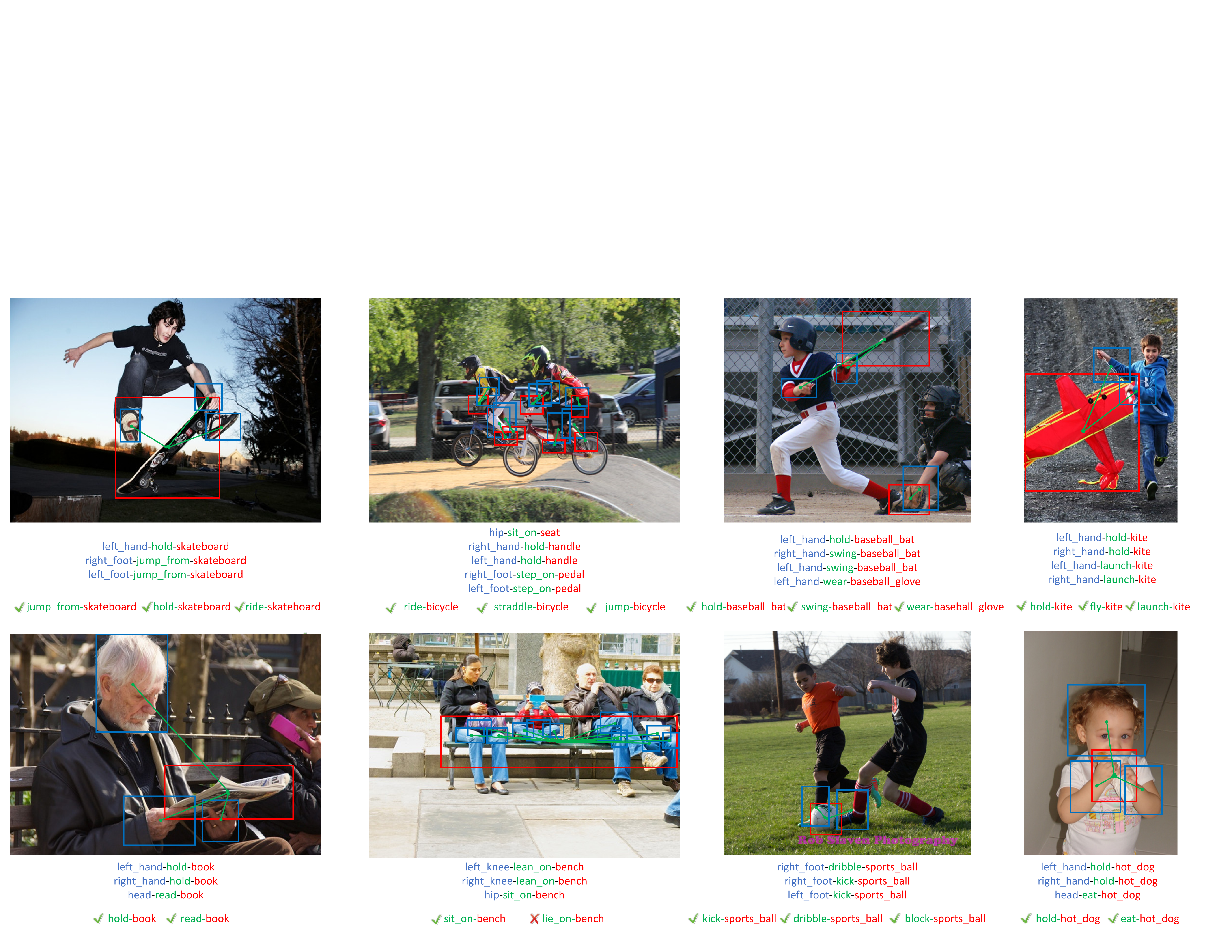}
	\end{center}
	\caption{Some predictions of our method. Triplets under images are predicted activities. Body part, part verb and object part are represented in blue, green and red, so are the activity results. Green tick means right prediction and red cross is the opposite.}
	\label{Figure:vis}
\end{figure*}
\begin{table}[H]
	\begin{center}
		\begin{tabular}{C{3.8cm}C{1.26cm}C{1.26cm}}
		\hline  
	    Method & mAP \\
	    \hline  
	    AlexNet+SVM~\cite{hico} & 19.4 \\ 
	    R*CNN~\cite{gkioxari2015contextual} & 28.5 \\
	    Girdhar \& Ramanan~\cite{girdhar2017attentional} & 34.6 \\
	    Mallya \& Lazebnik~\cite{Mallya2016Learning} & 36.1 \\
	    Pairwise~\cite{Fang2018Pairwise} & 39.9 \\
	    \hline
	    Pairwise~\cite{Fang2018Pairwise}+HAKE-GT & \textbf{62.5} \\
	    \hline
	    Pairwise~\cite{Fang2018Pairwise}+HAKE & \textbf{47.1} \\ 
	    \hline
	    Gain & \textbf{7.2} \\
	    \hline
		\end{tabular}
	\end{center}
	\caption{Comparison with previous methods on HICO. ``Pairwise~\cite{Fang2018Pairwise}+HAKE'' means the late fusion of the results from Pairwise~\cite{Fang2018Pairwise} and our HAKE.}
	\label{table:all}
\end{table}

\begin{table}[H]
	\begin{center}
		\begin{tabular}{C{3.8cm}C{1cm}C{1cm}C{1cm}C{1cm}}
		\hline  
	    Method & Few$@$1 & Few$@$5 & Few$@$10 \\
	    \hline  
	    Pairwise~\cite{Fang2018Pairwise} & 13.02 & 19.79 & 22.28 \\
	    Pairwise~\cite{Fang2018Pairwise}+HAKE & \textbf{25.40} & \textbf{32.48} & \textbf{33.71} \\
	    \hline
	    Gain & \textbf{12.38} & \textbf{12.69} & \textbf{11.43} \\
	    \hline
		\end{tabular}
	\end{center}
	\caption{Effectiveness on few-shot problems. Few$@$i represent the average mAP on few-shot activity sets. $@i$ means the number of training images is less than $i$, if $i$ is 1 then it means one-shot problem. On HICO~\cite{hico}, there is obvious positive correlation between performance and the quantity of training samples. Our approach can obviously improve the recognition effect on few-shot problem, for the reason of reusability and composability of part states.}
	\label{table:few}
\end{table}

From Tab.~\ref{table:all} we can find that our method achieve 7.2 mAP gain over the state-of-the-art result on HICO~\cite{hico}. 
And when the part state recognition is perfect, we can achieve surprising \textbf{62.5} mAP on HICO dataset. That is, if we input the ground truth part states into the Activity2Vec module and use the corresponding output activity representation vector for the classification, the \textbf{upper bound} of HAKE-based method is very high. This is a powerful proof of the representation ability and effectiveness of part states knowledge. Thus what remains to do for the activity understanding is to refine the part states recognition, then we can obtain more powerful activity representations and achieve better performance.

Moreover, on the one-shot and few-shot sets (training images are less than 5 and 10) of HICO, our method can achieve more than 11 mAP improvement. Specific results can be seen in Tab.~\ref{table:few}. These results show that our HAKE and HAKE-based hierarchical paradigm can significantly enhance the learning ability of model under few-shot circumstances.

Some qualitative results on HOI recognition are shown in Fig.\ref{Figure:vis}. The $\langle body\_part, part\_verb, object\_part\rangle$ with the highest scores are visualized in blue, green and red bounding boxes, and their corresponding labels are demonstrated under each image with colors consisted with boxes. The final predictions with the highest scores are represented too.

\section{Conclusion}
In this paper, we proposed a novel body part state knowledge base named Human Activity Knowledge Engine (HAKE) for human activity understanding, and a corresponding hierarchical paradigm. Our two-stage method consists of two components: Part State Recognition with Activity2Vec, and Part State based Reasoning.
With HAKE, we can obtain a new embedding combined both visual and linguistic semantic knowledge, which brings the interpretability in the human activity recognition. Part states can significantly heighten the activity reasoning to alleviate the learning difficulties brought by the imbalanced data, and utilize the co-occurrence relation to bridge the semantic interspace between part states and activity. Our experiment results on HOI recognition show that HAKE can significantly improve the performance, especially under the few-shot circumstances.

\section{Future Work}
Considering that our HAKE is still under construction, we will keep on enriching and enlarging it to promote the research in human activity understanding. We will also use HAKE to promote other tasks related to human activity understanding, \eg video-based activity understanding, action retrieval, Visual Question Answering and so on.

{\small
	\bibliographystyle{ieee}
	\bibliography{HAKE}
}

\clearpage
\begin{appendices}
Some samples from the proposed Part State Library, each part state consists of a state description and a corresponding cropped part region:
\begin{figure}[!ht]
	\begin{center}
		\includegraphics[width=0.45\textwidth]{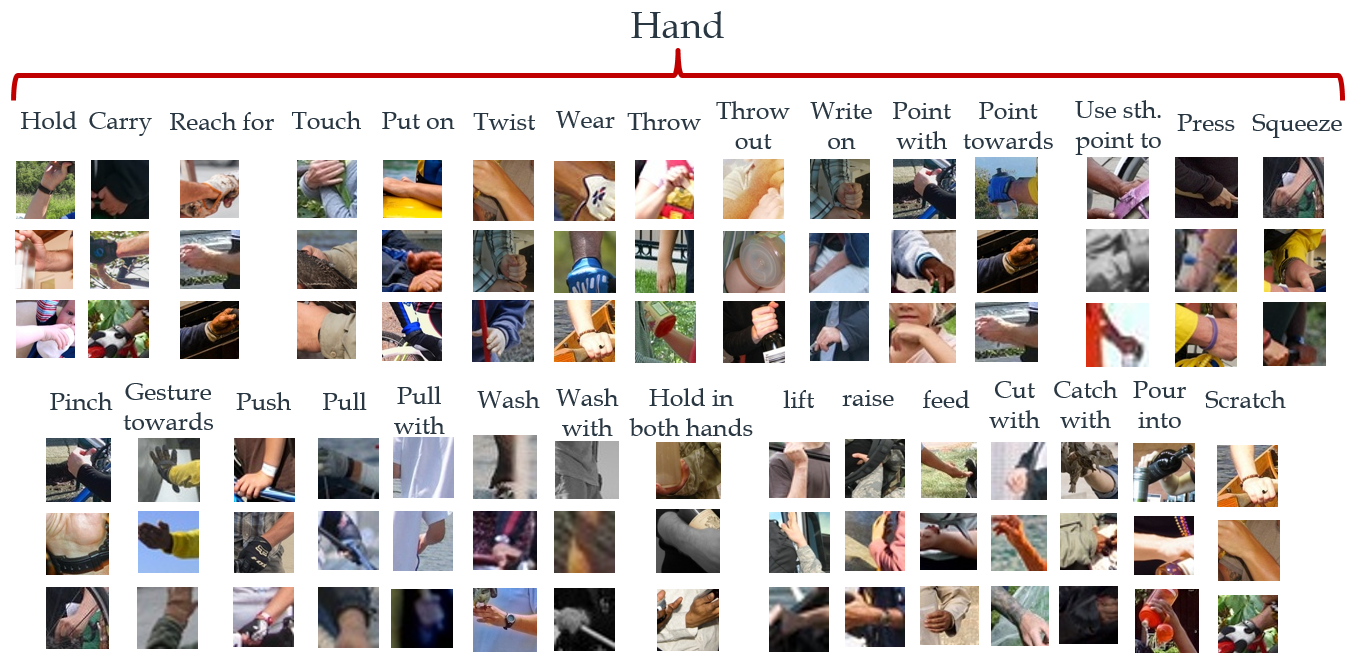}
	\end{center}
	\caption{Some ``hand'' states.}
	\label{Figure:ps_5}
\end{figure}
\begin{figure}[!ht]
	\begin{center}
		\includegraphics[width=0.45\textwidth]{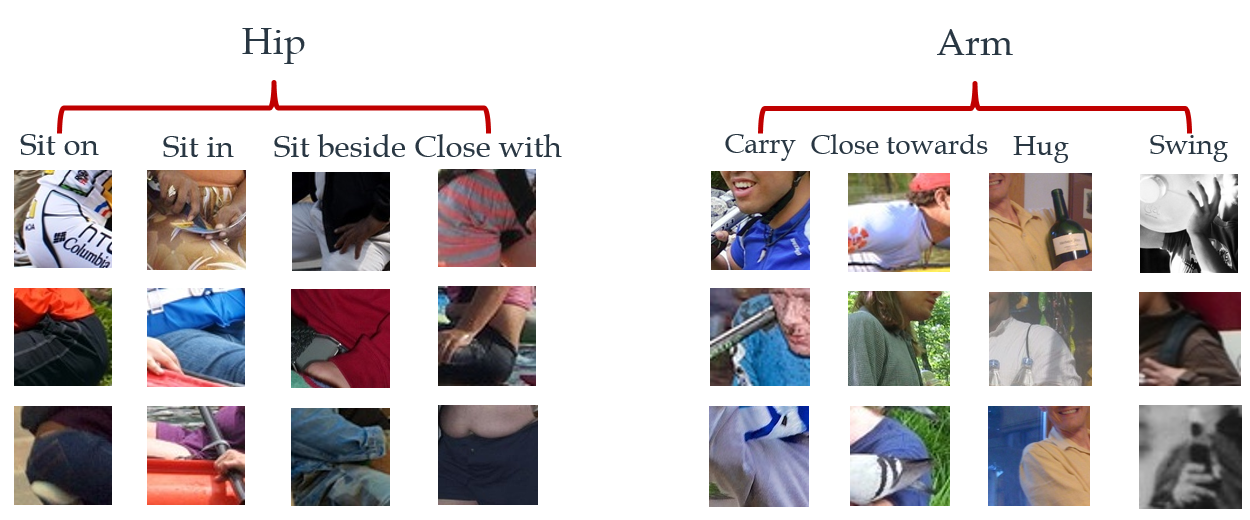}
	\end{center}
	\caption{Some ``hip'' and ``arm'' states.}
	\label{Figure:ps_1}
\end{figure}
\begin{figure}[!ht]
	\begin{center}
		\includegraphics[width=0.45\textwidth]{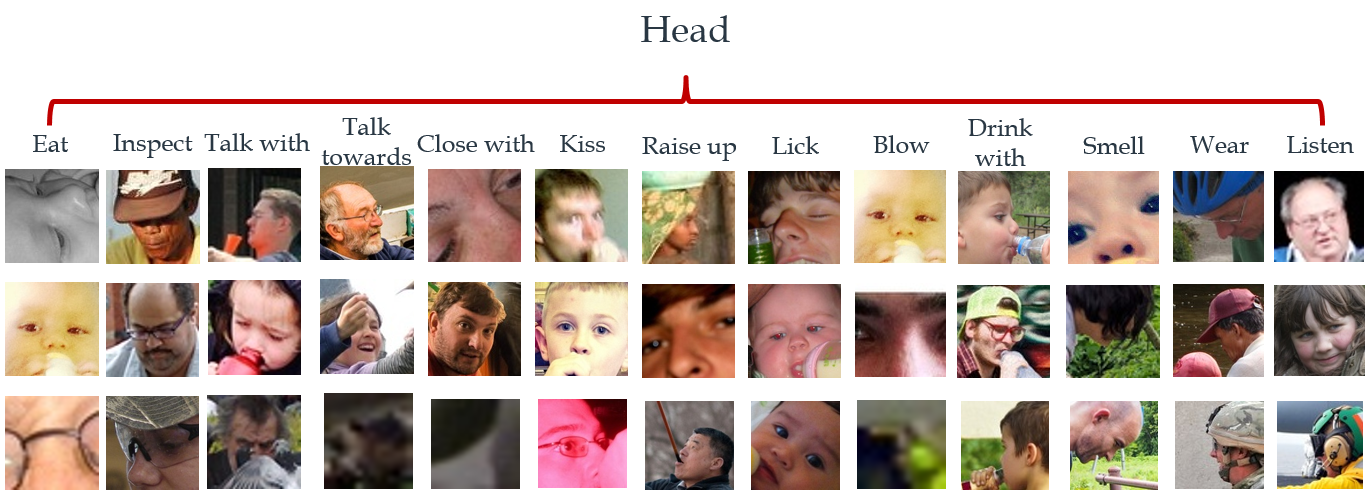}
	\end{center}
	\caption{Some ``head'' states.}
	\label{Figure:ps_2}
\end{figure}
\begin{figure}[!ht]
	\begin{center}
		\includegraphics[width=0.45\textwidth]{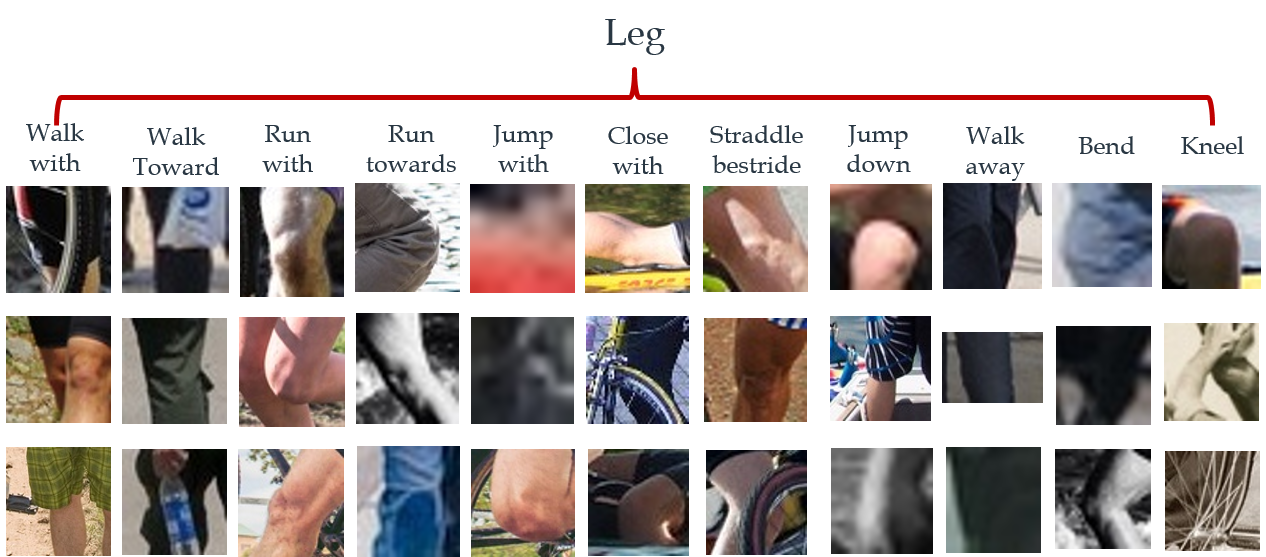}
	\end{center}
	\caption{Some ``leg'' states.}
	\label{Figure:ps_3}
\end{figure}
\begin{figure}[!ht]
	\begin{center}
		\includegraphics[width=0.45\textwidth]{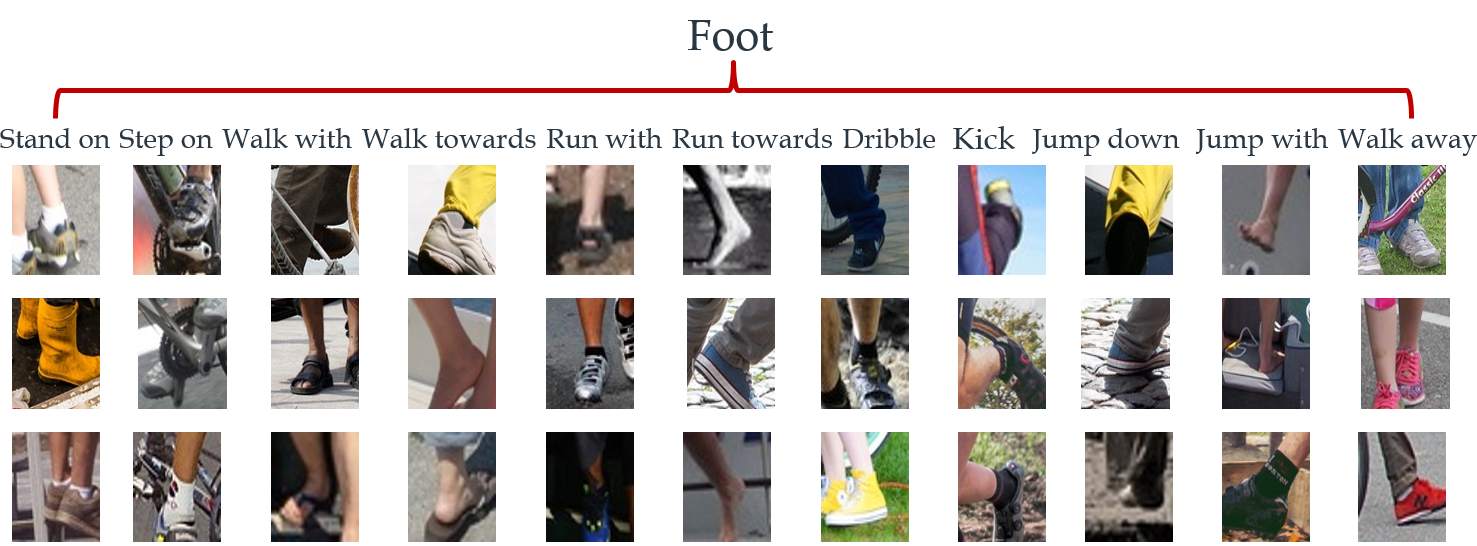}
	\end{center}
	\caption{Some ``foot'' states.}
	\label{Figure:ps_4}
\end{figure}

\end{appendices}
\end{document}